\documentclass{article}
\usepackage[nonatbib,preprint]{neurips_2020}
\usepackage[utf8]{inputenc} % allow utf-8 input
\usepackage[T1]{fontenc}    % use 8-bit T1 fonts
\usepackage{hyperref}       % hyperlinks
\usepackage{url}            % simple URL typesetting
\usepackage{booktabs}       % professional-quality tables
\usepackage{amsfonts}       % blackboard math symbols
\usepackage{nicefrac}       % compact symbols for 1/2, etc.
\usepackage{microtype}      % microtypography
\usepackage{xcolor}         % colors
\usepackage{adjustbox}
\usepackage{amsmath}
\usepackage{color}
\usepackage{algorithm}
\usepackage{algorithmic}
\usepackage{setspace}
\usepackage{wrapfig}
\usepackage{enumitem}
\usepackage{pifont} % for checkmark and cross
\setlist{topsep=0pt, leftmargin=*}
\usepackage{graphicx}
\usepackage{caption}
\usepackage{subcaption}
\usepackage{multirow}
\usepackage{makecell}

\usepackage[style=numeric,sorting=none]{biblatex}
\title{D3MES: Diffusion Transformer with multihead equivariant self-attention for 3D molecule generation}

\author{% 
Zhejun Zhang$^{1,2}$, \textbf{Yuanping Chen$^{1,2,\thanks{\scriptsize Corresponding author.}}$} \textbf{Shibing Chu$^{1,2,\footnotemark[1]}$}\\
\quad $^1$School of Physics and Electronic Engineering, Jiangsu University, \\Zhenjiang, Jiangsu, 212013, China. \\ 
\quad $^2$Jiangsu Engineering Research Center on Quantum Perception and Intelligent Detection \\of Agricultural Information, Zhenjiang, 212013, China.\\ 
\texttt{zhejunzhang0805@gmail.com}\\
\texttt{\{c, chenyp\}@ujs.edu.cn} \\
}

\begin{document}
\maketitle
\begin{abstract}
Understanding and predicting the diverse conformational states of molecules is crucial for advancing fields such as chemistry, material science, and drug development. Despite significant progress in generative models, accurately generating complex and biologically or material-relevant molecular structures remains a major challenge. In this work, we introduce a diffusion model for three-dimensional (3D) molecule generation that combines a classifiable diffusion model, Diffusion Transformer, with multihead equivariant self-attention. This method addresses two key challenges: correctly attaching hydrogen atoms in generated molecules through learning representations of molecules after hydrogen atoms are removed; and overcoming the limitations of existing models that cannot generate molecules across multiple classes simultaneously. The experimental results demonstrate that our model not only achieves state-of-the-art performance across several key metrics but also exhibits robustness and versatility, making it highly suitable for early-stage large-scale generation processes in molecular design, followed by validation and further screening to obtain molecules with specific properties. Code is available at https://github.com/PhysiLearn/D3MES.
\begin{figure}[htbp]
    \centering
    \includegraphics[width=0.7\textwidth]{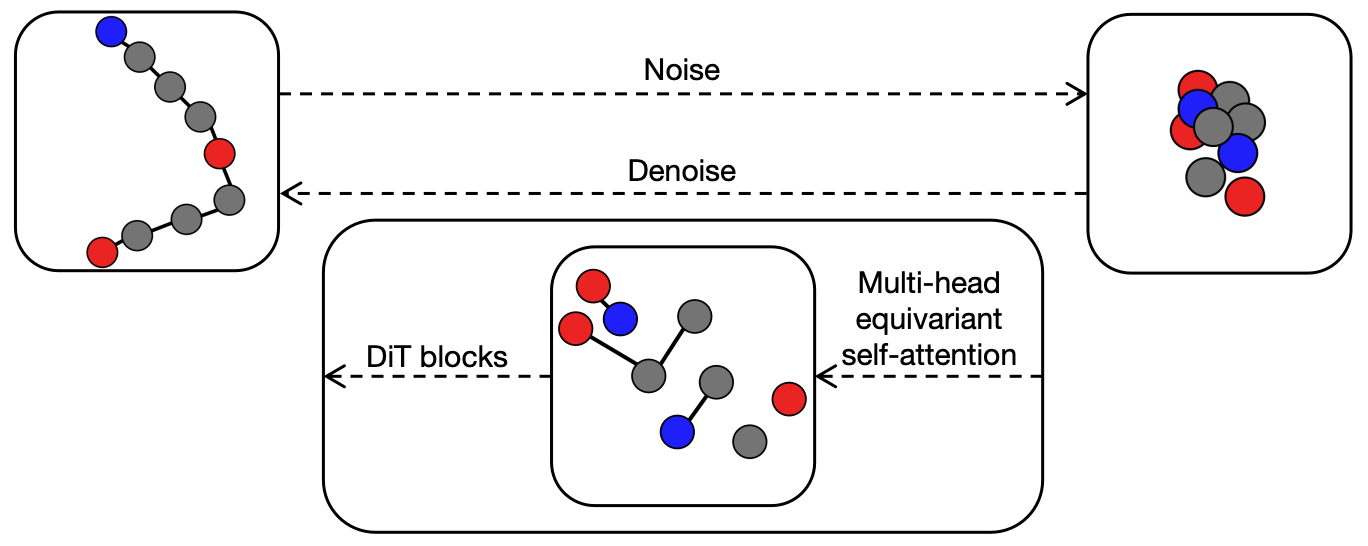} 
    \label{abstruct}
\end{figure}
\end{abstract}
\section{Introduction}
Molecular generation, especially ab initio 3D molecular generation, has become an important research focus in fields such as chemistry\cite{Huang2021}, material science\cite{Axelrod2022}, and drug development\cite{Xia2023}. Understanding the diverse conformational states of molecules plays a critical role in this process, as the biological or functional properties of a molecule are often highly dependent on its 3D conformation\cite{jumper2021highly}. Accurately predicting these conformations allows for the design of molecules with specific properties, thus accelerating discoveries in drug development, materials engineering, and other related areas. Traditional molecular generation methods rely primarily on high-throughput screening\cite{Gomez-Bombarelli2016}, which explores potential molecular candidates through extensive experiments and calculations. However, owing to the vast and complex chemical space\cite{Reymond2015,Ruddigkeit2012}, this approach not only requires substantial resources and time but also may fail to meet the needs for efficient screening and discovery owing to limitations in experimental conditions, screening accuracy, or strategies.

In contrast, ab initio 3D molecular generation aims to generate entirely new molecules with specific chemical or physical properties by learning the structure and properties of molecules through models\cite{Cho2019}. The advantage of this method lies in its ability to automatically explored and designed potential molecular structures from a molecular design perspective. By leveraging deep learning and generative models, these methods can generate molecules more accurately and efficiently and can handle and optimize complex 3D molecular structures, driving the design and application of drug molecules, materials, etc. Therefore, ab initio 3D molecular generation not only provides new ideas for scientific research but also offers more convenient and innovative solutions for applications such as drug development and material design\cite{Gupta2021,Hsu2022,zeni2023mattergen,li2024generative,runcie2023silvr}.

Early molecular generation methods focused primarily on two-dimensional (2D) representations. For example, SMILES expressions were used as input data, and recurrent neural networks (RNNs) \cite{ref7} were trained to generate molecular structures \cite{ref8,ref9}. While these methods perform well for simple molecules, they are limited in capturing the 3D spatial information of molecules, which is crucial for molecular property prediction. As the complexity of molecules increases, the limitations of 2D methods become more apparent, especially when dealing with large molecules, complex molecules, or drug molecules where the 3D structure plays a key role in determining their properties.

Recently, 3D molecular generation has gained significant attention. For example, G-Schnet \cite{ref18} combines an autoregressive process with the SchNet \cite{ref19} architecture, iteratively sampling atoms and bonds during generation. Similarly, E-NF \cite{ref20} adopts a one-shot framework to generate atom types and coordinates simultaneously. These 3D generation models offer new insights into drug design, as the 3D structure of molecules plays a vital role in determining their pharmacological activity, toxicity, and stability. To further enhance the generation of 3D molecular structures, diffusion models \cite{ref22} have been applied, which generate molecules through stepwise diffusion and denoising processes. For example, EDM \cite{ref23} integrates EGNN \cite{ref24} with a diffusion model to improve molecular generation precision and controllability.

However, despite the progress made with these models in 3D molecular generation, several challenges remain, particularly in determining the correct positioning of hydrogen atoms and generating molecules across multiple classes. Accurate placement of hydrogen atoms is critical for generating realistic drug molecules, but traditional 3D molecular generation methods often neglect this aspect. Moreover, existing models typically generate molecules from a single class and struggle to handle multiple classes of molecular data simultaneously, which is crucial for drug discovery, as drug molecules often contain diverse chemical categories and complex structures.

To address these challenges, this paper introduces a diffusion model that combines a classifiable diffusion model, Diffusion Transformer \cite{ref28}, with multihead equivariant self-attention \cite{ref33}, resulting in the Diffusion Transformer with multihead equivariant self-attention for molecular generation (D3MES). The D3MES model enhances the transformation of 3D molecular information after hydrogen atoms are removed, generating new feature representations through multihead equivariant self-attention. Additionally, D3MES is capable of simultaneously handling multiple classes of molecular data, overcoming the limitations of previous models.

We performed random molecular generation and classification-based molecular generation experiments on the small dataset GEOM-QM9\cite{ramakrishnan2014quantum} and evaluated random molecular generation on the larger dataset GEOM-Drugs\cite{axelrod2020geom}. The experimental results demonstrate that the D3MES model performs well across all the metrics, particularly standing out on the large molecule dataset.

By leveraging the D3MES model, candidate molecules can be efficiently generated in the early stages of molecular generation, significantly reducing the time required compared with traditional molecule synthesis methods. Through further validation and screening of these candidates, molecules with specific properties can be obtained.
\section{Theory}
\subsection{Diffusion Transformer}
The Diffusion Transformer (DiT) replaces the traditional convolutional U-Net\cite{ref29} backbone in the diffusion model with a transformer\cite{ref30} based architecture for learning and predicting noise $\epsilon_{\theta}$ and variance $\Sigma_{\theta}$. Specifically, DiT is built upon the Latent Diffusion Model (LDM)\cite{ref31}, which uses the Vision Transformer (ViT)\cite{ref32} as the backbone network and adapts the normalization techniques employed in ViT.

The input to DiT is the latent space $z = E(x)$, which is generated via VAE processing in the LDM, where $E(x)$ is the encoder of the VAE. This latent space $z = E(x)$ has dimensions $I \times I \times C$. The latent space $z$ is then divided into patches of size $P \times P$ through a patchify operation, forming $T = \left( \frac{I}{P} \right)^2$ sequences of tokens, each with dimensionality $d$.

For each token, the model applies Sinusoidal Positional Encoding (SPE), which is defined as: 
\begin{equation}
SPE(k, 2i) = \sin \left( \frac{k}{10000^{\frac{2i}{d}}} \right), \quad 
SPE(k, 2i+1) = \cos \left( \frac{k}{10000^{\frac{2i}{d}}} \right),
\end{equation}
where \( k \) represents the position index of the token in the sequence and \( i \) is the dimension index of the positional encoding. This encoding helps the model incorporate the order of tokens in the sequence by using sine and cosine functions with different frequencies. After this step, the position-encoded tokens are passed into the transformer network for further processing. Within the transformer, each token \( z_t \) is processed through a self-attention layer, where the attention score \( A \) is computed via the following formula:
\begin{equation}
A = \text{softmax} \left( \frac{QK^T}{\sqrt{d_k}} \right).
\end{equation}
Here, \( Q \) represents the query matrix, which encodes the query information for each token in the sequence, and \( K \) represents the key matrix, which is used to compute attention scores by comparing it with the query matrix. The denominator \( \sqrt{d_k} \) is used to scale the dot product, where \( d_k \) is the dimension of the key vectors. This scaling factor helps prevent excessively large dot product values, which could lead to numerical instability in the Softmax operation. Once the attention scores are calculated, they are used to compute a weighted sum of the input features, which allows the model to focus on the most relevant tokens in the sequence.

After passing through the DiT blocks, the model outputs the noise prediction \( \epsilon_{\theta} \) and the variance prediction \( \Sigma_{\theta} \). The training loss for the noise prediction network is computed via the Mean Squared Error (MSE) loss:
\begin{equation}
L_{\text{simole}} (\theta) = \|\epsilon_{\theta} (x_t) - \epsilon_t \|^2.
\end{equation}
In this equation, \( \epsilon_t \) represents the true noise sampled from a standard Gaussian distribution \( N(0,1) \), and \( x_t \) respresents the sample data with added noise, which serves as the noisy input for training. The model’s goal is to minimize the difference between the predicted noise and the true noise, thereby learning to recover the original clean data from the noisy input during training.
\subsection{Multihead Equivariant Self Attention in the SE3-Transformer}
The SE(3)-Transformer\cite{ref33} is a variant of the transformer model designed for tasks involving 3D rotations and translations. This ensures spatial invariance, meaning that the output remains consistent regardless of how the input is rotated or translated.

The attention mechanism in this model is designed to maintain consistency under SE(3) transformations, meaning that the attention weights remain unchanged when the data are rotated or translated. This allows the model to effectively process 3D data without being affected by its position or orientation in space.

Given a point cloud $\{(x_i, f_i)\}$, denote the neighborhoods as $N_i \subseteq \{1, \dots, N\}$, each centered around $i$. The attention to each of the neighborhoods is executed on the basis of the following:
\begin{equation}
f_{\text{out},i}^l = W_V^{l} f_{\text{in},i}^l + \sum_{k \geq 0} \sum_{j \in N_i \setminus i} \alpha_{ij} W_V^{lk} (X_j - X_i) f_{\text{in},j}^k.
\label{eqn:out_i}
\end{equation}

The first part of the equation involves applying a linear transformation to the input features \(f_{\text{in},i}^l\) of point \(i\) via the weight matrix \(W_V^{l}\), which results in a linear transformation of the point's own features. The second part involves the weighted sum of the features from the neighboring points \(N_i \setminus i\), where the weights are determined by the attention coefficient \(\alpha_{ij}\). The weighted features are the product of the relative position difference \((x_j - x_i)\) between point \(i\) and its neighbor point \(j\), and the input features \(f_{\text{in},j}^k\) of the neighboring point \(j\). The introduction of the attention mechanism enables the model to update the output features of each point on the basis of the weighted contributions from its neighboring points' features,
where the definitions for $\alpha_{ij}$ and the parameters $q_i$ and $k_{ij}$ are shown below:
\begin{equation}
\alpha_{ij} = \frac{\exp(q_i^\top k_{ij})}{\sum_{j' \in N_i \setminus i} \exp(q_i^\top k_{ij'})}, \quad
q_i = \bigoplus_{l \geq 0} \sum_{k \geq 0} W_Q^{lk} f_{\text{in},i}^k, \quad
k_{ij} = \bigoplus_{l \geq 0} \sum_{k \geq 0} W_k^{lk} (x_j - x_i) f_{\text{in},j}^k.
\end{equation}

Here, $\oplus$ denotes the direct sum, and the linear embedding matrices $W_Q^{lk}$ and $W_k^{lk}(x_j - x_i)$ are of the tensor field network\cite{Thomas2018TensorFieldNetworks} type, satisfying the following equation:
\begin{equation}
W^{lk}(x) = \sum_{J = |k - l|}^{k + 1} \phi_J^{lk} (\|x\|) W_J^{lk}(x), \quad
W_J^{lk}(x) = \sum_{m = -J}^{J} Y_{Jm} \left(\frac{x}{\|x\|} \right) Q_{Jm}^{lk}.
\label{eqn:W_lk_x}
\end{equation}

First, \(W^{lk}(x)\) is the weight matrix at layer \(l\) and neighborhood level \(k\), adjusted on the basis of the distance \(\|x\|\) between points, allowing the network to adaptively transform features according to geometric relationships. \(\phi_J^{lk}(\|x\|)\) modulates the influence of each neighborhood level depending on the distance \(\|x\|\). \(W_J^{lk}(x)\) incorporates both the radial distance and the directional information, with \(x = x_j - x_i\) representing the relative position of point \(j\) to point \(i\). The spherical harmonic function \(Y_{Jm} \left(\frac{x}{\|x\|}\right)\) captures the directional relationship between neighboring points, whereas \(Q_{Jm}^{lk}\) is a learned coefficient matrix that adjusts the impact of each directional component, enabling the model to focus on specific spatial directions and improve feature transformation accuracy.

If the input features $f_{\text{in},j}$ are SO(3) equivariant, then both the query vector $q_i$ and the key vector $k_{ij}$ will also be equivariant, as they are obtained through equivariant linear embedding matrices. Moreover, the orthogonality of the SO(3) group representation ensures the SO(3) invariance of the dot product of these vectors:
\begin{equation}
q \mapsto S_g q, \quad k \mapsto S_g k \quad \Rightarrow \quad q^\top S_g^\top S_g k = q^\top k.
\label{eqn:transform_eq}
\end{equation}

Thus, $S_g$ is the transformation matrix of the SO(3) group, which represents a rotation operation.
The attention weights $\alpha_{ij}$ are invariant under SO(3) transformations, ensuring that the entire attention mechanism is SE(3) equivariant.
\section{Methods}

We add a layer of multihead equivariant self-attention\cite{ref33} on the basis of Diffusion Transformer\cite{ref28}. The addition of the attention mechanism can better strengthen the molecular features, making the subsequent network processing more efficient.
\subsection{Data Preprocessing}

\begin{figure}[htbp]
    \centering
    \includegraphics[width=0.7\textwidth]{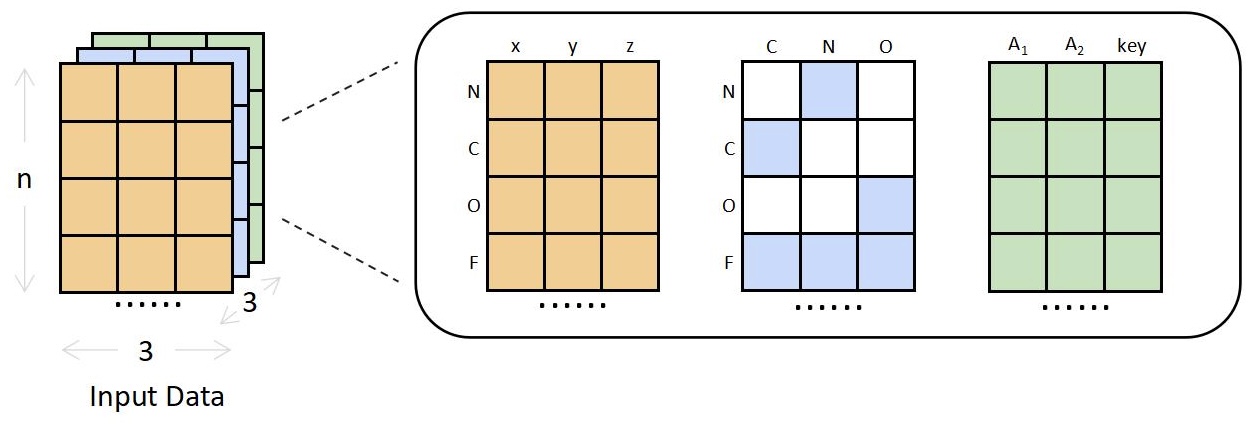}  % 图片路径为1.jpg，确保图片与.tex文件在同一文件夹，或者提供正确的路径
    \caption{Data preprocessing. The molecular information is deposited into three channels, the first channel is the 3D coordinates of the atoms in the molecule, the second channel is the elemental information of the molecule, and the third channel is the bond connectivity information of the molecule.}
    \label{fig:data_preprocessing}
\end{figure}

We transform each set of input data into 3 channels, where each set of data represents one molecule. The first channel stores the position of each atom in the molecule, the second channel stores the elemental information of each atom, and the third channel represents the bonding information between atoms in the molecule, as shown in Figure \ref{fig:data_preprocessing}.

% 插入图像

The reason for this characterization of the data is to better incorporate multihead equivariant self-attention, which allows for easier processing of the data. Once the data preprocessing is complete and noise is added, the second channel's data are used as the node features for the attention mechanism, whereas the key-connection information from the third channel serves as the edge features.

We extract each nonhydrogen atom in the molecule as a point cloud \(\{(x_i, f_i)\}\) and denote the neighborhoods as \(N_i \subseteq \{1, \dots, N\}\), each centered around point \(i\). The query and key vectors are computed from the node and edge features via equation (5), where \((x_j - x_i)\) represents the distance between atoms, which is subsequently used to compute the attention weights. The new feature representation is obtained via equation (4).

\begin{figure}[htbp]
    \centering
    \includegraphics[width=0.7\textwidth]{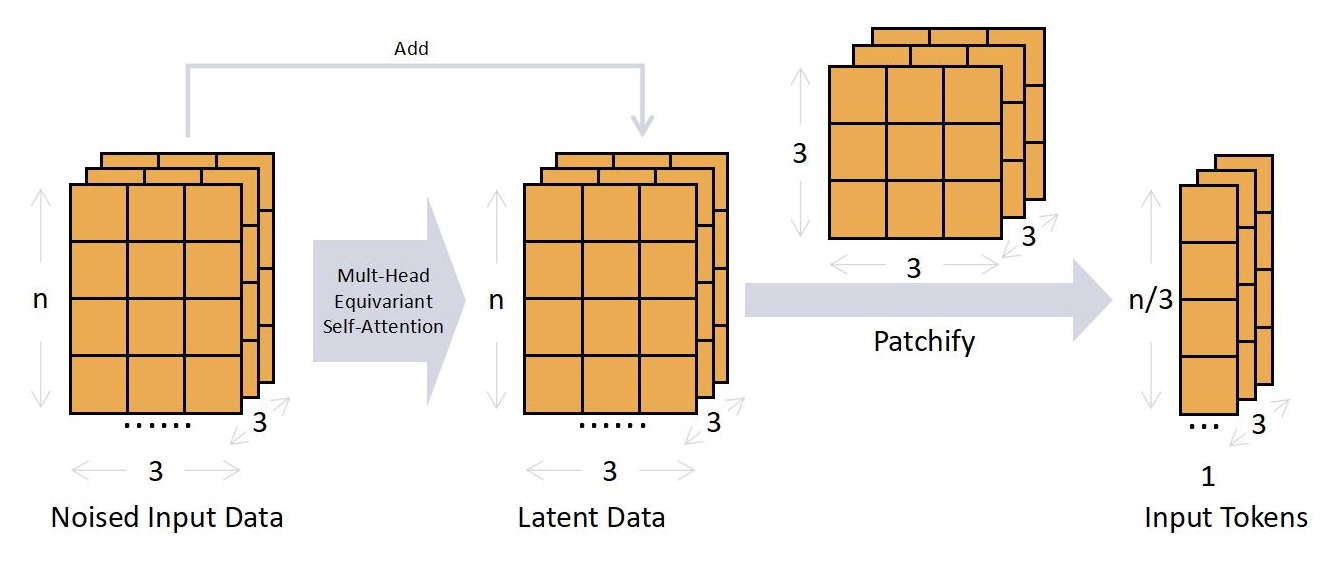}
    \caption{Preprocessing of data into DiT blocks. Input data are transformed into new feature data by the attention mechanism, which is then combined with the original input data and passed through the patchify process to generate tokens.}
    \label{fig:equattention}
\end{figure}

After calculating the attention weights, we replace the information in the first channel of the input data. After the replacement is completed, the data are processed into $3 \times 3$ patch panels and further processed as input to the DiT blocks, as shown in Figure \ref{fig:equattention}.
\subsection{The Training Process}

This approach enables the geometric structure of the input data to be more efficiently encoded through the equivariant attention mechanism, thereby enhancing the model's feature extraction capabilities.

\begin{figure}[htbp]
    \centering
    \includegraphics[width=0.7\textwidth]{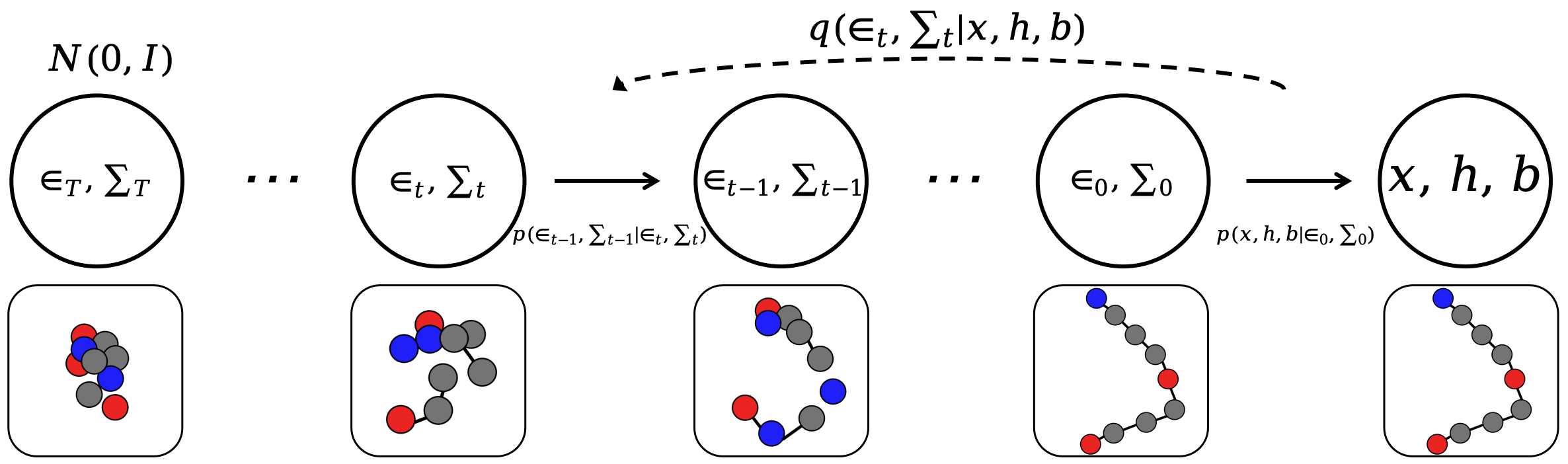}  
    \caption{Overview of the diffusion process. To generate the molecule, the process begins by initializing the noise $\epsilon_T$ and variance $\Sigma_T$, followed by iterative denoising of these variables. The goal is to generate the coordinates $x$, elemental information $h$, and bond connectivity information $b$. This is achieved by sampling from the distribution $p(\epsilon_{t-1}, \Sigma_{t-1} |\epsilon_t, \Sigma_t)$ at each iteration. During training, the model $q(\epsilon_t, \Sigma_t | x, h, b)$ at time step $t$ is used to add noise and variance to the data points $x$, $h$, $b$ to learn the denoising process.}
    \label{fig:diffusion_process}
\end{figure}

Suppose there is a set of real molecular data $x_0$, which contains atomic position information, elemental information, and bond connectivity information in the molecule. and noise is progressively introduced through a forward noise addition process, resulting in $x_T$, which eventually converges to standard Gaussian noise. This process of noise addition in the forward direction is as follows:
\begin{equation}
    q(x_t | x_{t-1}) = \mathcal{N}(x_t; \sqrt{\alpha_t} x_{t-1}, (1-\alpha_t)I).
    \label{eqn:q_x_t}
\end{equation}

Here, $x_t$ represents the data at time step $t$, and $\alpha_t$ is a hyperparameter that governs the noise intensity. This parameter typically increases as $t$ increases, thereby increasing the proportion of noise over time.
 $I$ is the identity matrix, which represents the variance of the noise addition process. After $T$ iterations, $x_T$ will eventually become a standard normal distribution noise: $x_T \sim \mathcal{N}(0, I)$.

During training, the goal is to learn the inverse diffusion process $p_\theta(x_{t-1} | x_t)$, which aims to recover the original data from the noisy observations. The formula for the inverse process is as follows:
\begin{equation}
    p_\theta(x_{t-1} | x_t) = \mathcal{N}(x_{t-1}; \mu_\theta(x_t), \Sigma_\theta(x_t)).
    \label{eqn:p_theta}
\end{equation}

In this context, $\mu_\theta(x_t)$ represents the noise prediction of the model, and $\Sigma_\theta(x_t)$ denotes the variance prediction of the model. The model is defined by $\mu_\theta(x_t)$ and $\Sigma_\theta(x_t)$. The goal of the inverse denoising process is to minimize the error between the predicted noise and the true noise.

Notably, in noise prediction, the model must learn to predict noise on the basis of the given input $x_t$ and additional conditions. In addition to predicting the noise, the model also needs to predict the noise variance at each time step and optimize the predicted variance via the Kullback-Leibler (KL) divergence loss, as follows:
\begin{equation}
    L(\theta) = -\log p(x_0 | x_1) + \sum_t D_{\text{KL}}(q^*(x_{t-1} | x_t, x_0) \parallel p_\theta(x_{t-1} | x_t)).
\label{eqn:loss_function}
\end{equation}

This process aims to minimize the difference between the noise and variance predicted by the model and the true data. Specifically, $q^*(x_{t-1} | x_t, x_0)$ represents the true inverse process when $x_t$ and $x_0$ are known, whereas $p_\theta(x_{t-1} | x_t)$ is the inverse process predicted by the model. By minimizing this loss, the model learns to predict not only the noise but also the variance of the noise.

\begin{figure}[htbp]
    \centering
    \includegraphics[width=0.7\textwidth]{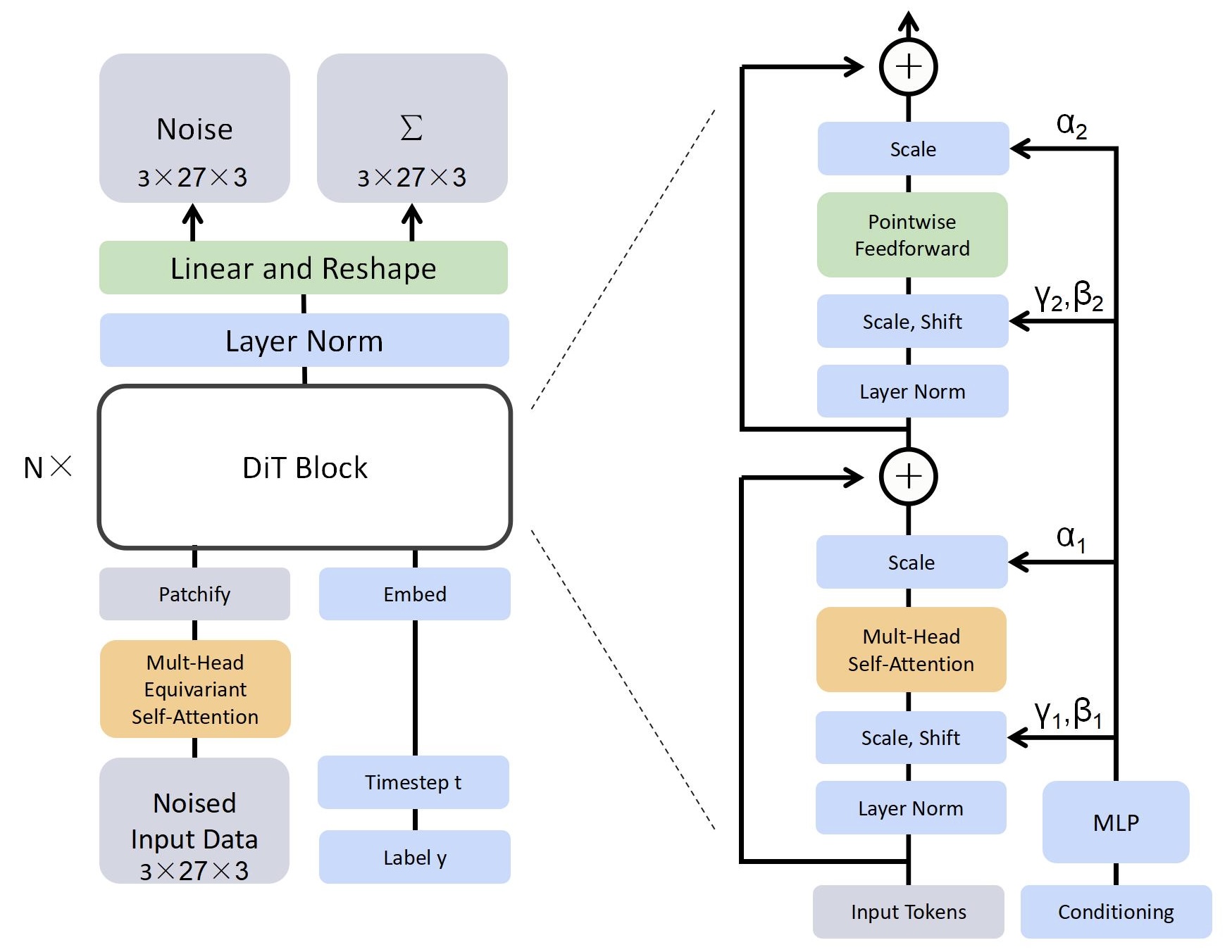}  % 图片路径为4.jpg，确保图片与.tex文件在同一文件夹，或者提供正确的路径
    \caption{D3MES. The normalized block version of the DiT block with adaLN-Zero, with an additional layer of multihead equivariant self-attention.}
    \label{fig:dtmeam_flowchart}
\end{figure}

Building on the above discussion, our model integrates the DiT block with adaLN-Zero and multihead equivariant self-attention. The overall process is illustrated in Figure \ref{fig:dtmeam_flowchart}.

In our approach, we discard the encoder and decoder components of the VAE and introduce an additional layer of attention mechanism in front of the DiT blocks. This enhancement improves the model’s ability to capture features at different scales and orientations, enabling it to better learn the feature representations of the input data.

After the feature representations are obtained, the data undergo a patching process. The input data are split into patches of size $p \times p$ via the patchify operation, resulting in a sequence of tokens $T = \left(\frac{h}{p} \right) \left( \frac{w}{p} \right)$, where $h$ and $w$ represent the height and width of the input data. Following these transformations, the data are passed into the DiT blocks for further processing.

In particular, after the final DiT block, the model linearly decodes each token into a tensor of size $p \times p \times 2c$, where the first $c$ corresponds to the noise prediction $\hat{\epsilon}_\theta(x_t, c)$, and the second $c$ corresponds to the variance prediction $\hat{\Sigma}_\theta(x_t, c)$. Both predictions have the same spatial layout as the input data, resulting in noise and variance predictions at each time step.

\subsection{Generation}

In the generation process, the model begins with standard Gaussian noise $x_T \sim \mathcal{N}(0, 1)$ and progressively denoises it to recover the original data $x_0$, by leveraging the noise prediction $\hat{\epsilon}_\theta(x_t, c)$ and the variance prediction $\hat{\Sigma}_\theta(x_t, c)$ learned during training. This process is described by the following equation:
\begin{equation}
    x_{t-1} = \frac{1}{\sqrt{\alpha_t}} \left( x_t - \frac{1-\alpha_t}{\sqrt{1-\bar{\alpha}_t}} \hat{\epsilon}_\theta(x_t, c) \right) + \sqrt{\Sigma_{t^\epsilon}},
    \label{eqn:formula_15}
\end{equation}

where $\epsilon$ is sampled from the standardized Gaussian distribution. In this way, the model gradually removes noise through a process of reverse diffusion, ultimately generating new data that contain atomic position information, elemental information, and bond connectivity information in the molecule.

\section{Experiments}
\begin{figure}[htbp]
    \centering
    \includegraphics[width=\textwidth]{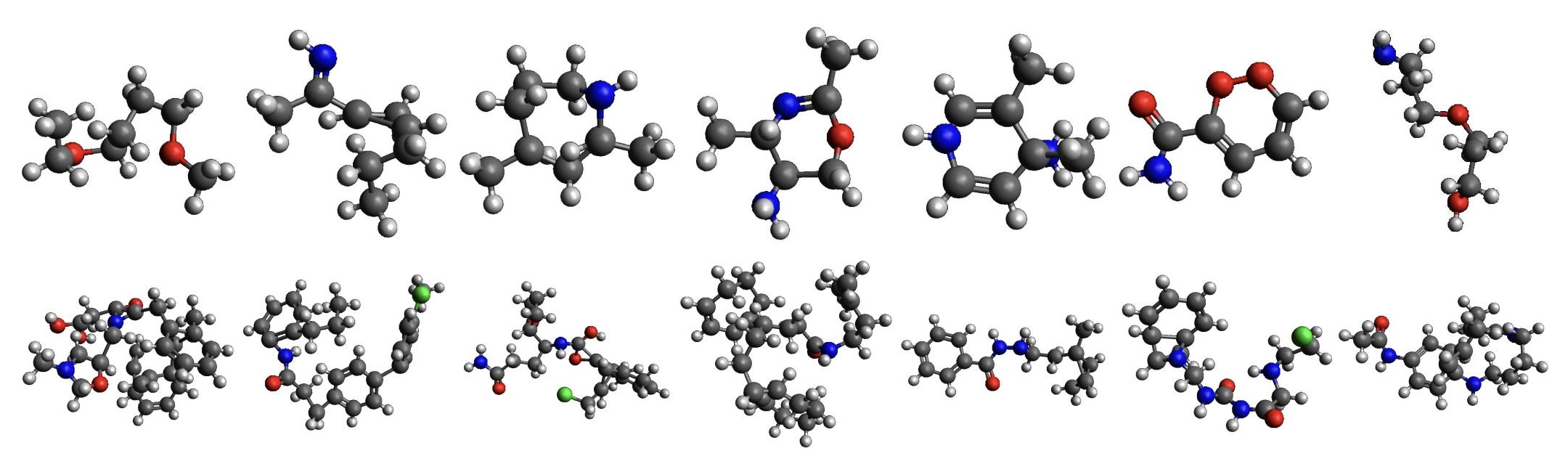}  % 图片路径为8.jpg，确保图片与.tex文件在同一文件夹，或者提供正确的路径
    \caption{Random generation of molecules. The molecules above are generated on the basis of the QM9 dataset, and the molecules below are generated on the basis of the Drugs dataset.}
    \label{fig:qm9_generation}
\end{figure}

\subsection{Molecular Generation - QM9}

GEOM-QM9 is a standard dataset containing the molecular properties and atomic coordinates of 130k small molecules, each with up to 9 heavy atoms (29 atoms in total, including hydrogen). In this experiment, we trained D3MES to unconditionally generate molecules, including their 3D coordinates, atom types (C, N, O, F), and interatomic bond connections. To demonstrate the effectiveness of the multihead equivariant self-attention mechanism in our model, we also trained a version of the model without the attention, referred to as the DTM.

\noindent \textbf{Metrics:} For the key connections between atoms, we adopt the same approach as EDM, inferring them via element information and atomic positions. Then, the atomic stability can be evaluated by measuring the proportion of atoms with the correct valence, and the molecular stability can be evaluated by calculating the proportion of molecules in which all the atoms are stable. Additionally, we evaluate the validity of the generated molecules by using RDKit and assess the uniqueness of the resulting compounds by calculating the proportion of unique generated SMILES.

\noindent \textbf{Baselines:} We compare DTM and D3MES with existing molecular generative models that use diffusion as their base model: EDM\cite{ref23}, GCDM\cite{ref34}, MDM\cite{ref25}, JODO\cite{ref35}, MiDi\cite{ref26}, and GeoLDM\cite{ref27}. For each generative model, we use the best results reported in the corresponding paper for comparison.

\begin{table}[htbp]
  \centering
  \begin{tabular}{lllll} 
    \hline
    Model & At Stable(\%) & Mol Stable(\%) & Valid(\%) & Val/Uniq(\%) \\
    \hline
    EDM & 98.7 & 82.0 & 91.9 & 90.7 \\
    GCDM & 98.7 & 85.7 & 94.8 & 93.3 \\
    MDM & 99.2 & 89.6 & 98.6 & 94.6 \\
    JODO & 99.2 & 93.4 & \textbf{99.0} & 96.0 \\
    MiDi & \textbf{99.8} & \textbf{97.5} & 97.9 & 97.6 \\
    GeoLDM & 98.9 & 89.4 & 93.8 & 92.7 \\
    DTM & 99.0 & 92.2 & 97.6 & 98.1 \\
    D3MES(ours) & 99.3 & 92.2 & 97.8 & \textbf{98.3} \\
    \hline
  \end{tabular}
  \caption{Comparison of atomic stability, molecular stability, validity, and uniqueness. The DTM and D3MES models are trained on the QM9 dataset. A total of 10,000 samples are generated from each model, and the atomic stability, molecular stability, validity, and uniqueness are computed and compared with those of other models.}
  \label{tab:qm9_comparison}
\end{table}

\noindent \textbf{Results:} The results on the QM9 dataset are shown in Table \ref{tab:qm9_comparison}, where our model achieves excellent performance across the four metrics. Specifically, our model scores 99.3\% on the atom stable metric, ranking second among the eight models compared. For the metric of molecular stability, our model ranks third with a commendable result. For the valid measurement, our model ranks fourth but is only 0.1\% behind the MiDi model in third place. Notably, our model achieves the best performance in the uniqueness measurement, with an impact score of 98.3\%.
In the comparison between DTM and D3MES, the model enhanced with multihead equivariant self-attention outperforms the baseline model across almost all the metrics. This improvement is attributed primarily to the ability of attention mechanisms to integrate the structural and property information of molecules more comprehensively. Through this mechanism, the model can more accurately capture critical molecular features, enabling more efficient data processing.

\subsection{Molecular Generation - Drugs}

To demonstrate the generalizability of the model, we trained the model on the large molecule dataset GEOM-Drugs. This dataset contains more than 290k molecules, each with up to 91 heavy atoms (181 atoms in total, including hydrogen). 
For training, we selected the 30 lowest-energy conformations for each molecule.

\noindent \textbf{Baselines:} We compare D3MES with existing molecular generative models that use diffusion as their base model: EDM, MDM, MiDi, and GeoLDM. For each generative model, we use the best results reported in the corresponding paper for comparison.

\begin{table}[htbp]
\centering
\begin{tabular}{lllll}
\hline
Model & At Stable(\%) & Mol Stable(\%) & Valid(\%) & Val/Uniq(\%) \\
\hline
EDM & 81.3 & / & / & / \\
MDM & / & 62.2 & 99.5 & 99.0 \\
MiDi & \textbf{99.8} & 91.6 & 77.8 & 77.8 \\
GeoLDM & 84.4 & / & 99.3 & / \\
D3MES(ours) & \textbf{99.8} & \textbf{94.7} & \textbf{99.98} & \textbf{99.9} \\
\hline
\end{tabular}
\caption{Comparison of atomic stability, molecular stability, validity, and uniqueness. D3MES is trained on the Drugs dataset. A total of 10000 samples are generated from this model, and the atomic stability, molecular stability, validity, and uniqueness are computed and compared with those of other models.}
\label{tab:drugs_comparison}
\end{table}

\noindent \textbf{Results:} The results on the Drugs dataset are shown in Table \ref{tab:drugs_comparison}, where our model achieves excellent performance across all the metrics. For the Atom stable metric, our model achieves the same result as MiDi, which ranks first. For the molecular stability metric, our model outperforms all others by a margin of 3.1\%, indicating the top position. For the valid metric, our model achieves an impressive 99.98\%, meaning that out of 10,000 generated molecules, only two failed from RDKit's validation. Additionally, our model achieves the best result for the uniqueness metric, reaching 99.9\%.
Since the base model we selected uses the patchify layer, which effectively divides and consolidates information from large datasets, is used to process the data. As a result, our D3MES performs exceptionally well on this dataset, which contains larger and more structurally complex molecules.
\subsection{Molecular Classification Generation}

\begin{figure}[htbp]
    \centering
    \includegraphics[width=\textwidth]{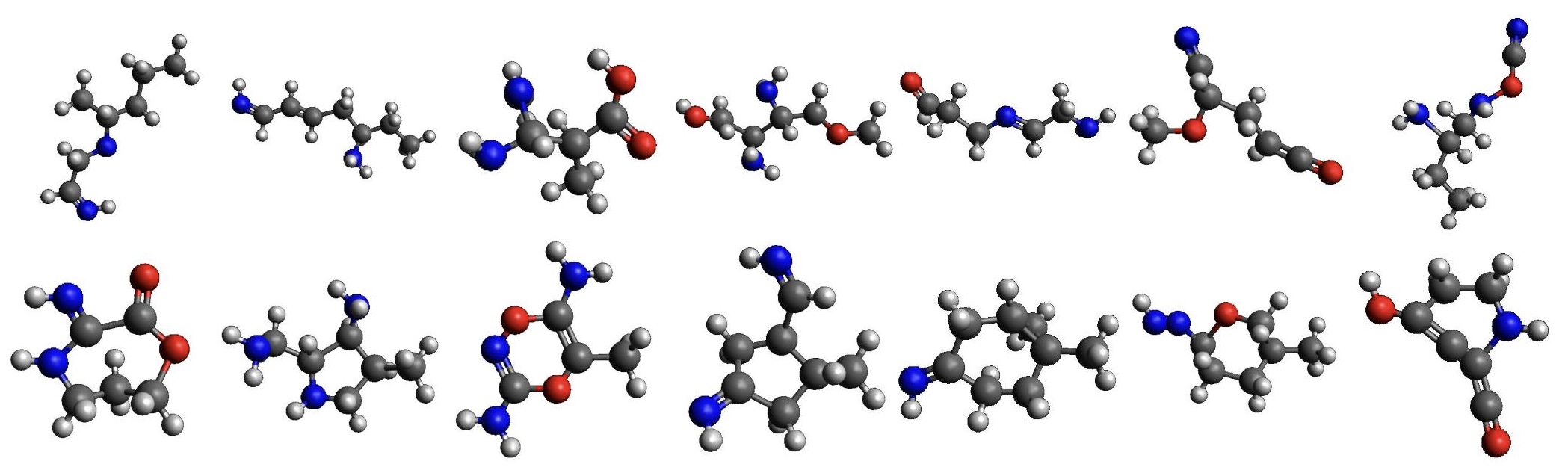}  
    \caption{Molecular classification generation based on the QM9 dataset. The top section displays noncyclic molecules, and the bottom section shows cyclic molecules.}
    \label{fig:classification_generation}
\end{figure}

\noindent \textbf{Metrics:} In this section, we extract both noncyclic and cyclic molecules from the QM9 dataset to form a subset of approximately 30k molecules. After training on this subset, we generate 1,000 molecules for each category (noncyclic and cyclic) via all the corresponding channel data. We then verify each generated molecule for the presence of cyclic structures and calculate the accuracy of the generated molecules for each category.

\noindent \textbf{Results:} After training our model on a subset of approximately 30,000 molecules extracted from the QM9 dataset, we generated 1,000 molecules for each category. 
The generated molecules were then verified for the presence of cyclic structures, and the accuracy of each category was calculated. The results show that our model achieved an accuracy of \textbf{97.4\%} for noncyclic molecules and \textbf{96.6\%} for cyclic molecules. These results demonstrate the model's ability to generate high-quality molecules that are well-aligned with their designated categories. 
The small difference in accuracy between noncyclic and cyclic molecules further demonstrates the stability and adaptability of our model in handling different molecular structures.

\section{Discussion and Conclusions}

We introduce D3MES, a classifiable diffusion model with multihead equivariant self-attention, which updates molecular feature information more efficiently through the attention mechanism. The model performs excellently on the small molecule dataset QM9 and achieves the best results in three metrics on the large molecule dataset Drugs. By applying patchification to modularize the data, the model efficiently handles larger datasets, making it highly effective for generating large molecules. Its unique architecture allows for the simultaneous processing of multiple types of molecular information, which is crucial in the field of molecular generation. In chemistry, material science, or drug development, this approach accelerates the early-stage generation of potential candidate molecules and lays a solid foundation for future advancements in molecular classification and generation. The current method could be improved in several ways, but these improvements are beyond the scope of this work.

\textbf{Optimization of Hydrogen Atom Addition}. Currently, in the molecular generation process, hydrogen atoms are added on the basis of the maximum allowable valence state for each atom. This approach may overlook other possible configurations, leading to suboptimal molecular structures. A potential future direction is to introduce a dedicated training module for hydrogen atom addition during the generation process. This would enable more precise handling of hydrogen atom placement, taking into account the specific context of the molecule being generated.

\textbf{Performance differences on the small- and large-molecule datasets}. Although the patchification module enhances the efficiency of processing large datasets, its performance on small datasets (such as QM9) is relatively less optimal. A potential future direction is to implement multiple training pathways tailored to datasets of different scales. For smaller datasets, lightweight modules could be employed, whereas for larger datasets, the full patchification architecture can be utilized, ensuring optimal performance across datasets of various sizes.
\section*{Acknowledgement}
This work gratefully acknowledges the National Natural Science Foundation of China (No. 11904137, 12074150 and 12174157)

{\small
\printbibliography
}

\end{document}